\documentclass{article}

\PassOptionsToPackage{numbers, sort&compress}{natbib}

\usepackage[preprint]{neurips_2020}

\usepackage[utf8]{inputenc} 
\usepackage[T1]{fontenc}    
\usepackage{hyperref}       
\usepackage{url}            
\usepackage{booktabs}       
\usepackage{amsfonts}       
\usepackage{nicefrac}       
\usepackage{microtype}      

\usepackage[utf8]{inputenc}
\usepackage{amsmath,amsfonts,amssymb,amsthm}
\usepackage{xcolor}
\usepackage{booktabs} 
\newtheorem{theorem}{Theorem}
\newtheorem{lemma}[theorem]{Lemma}
\newtheorem{assumption}[theorem]{Assumption}
\usepackage{bm}
\usepackage{graphicx}
\usepackage{wrapfig}

\newcommand{\vv}{\mathrm{v}}
\newcommand{\vs}{\mathrm{s}}
\newcommand{\vg}{\mathrm{g}}

\newcommand{\ls}{\mathcal{L}}
\newcommand{\lshat}{\hat{\mathcal{L}}}

\newcommand{\eg}{\textit{e.g.}}
\newcommand{\ie}{\textit{i.e.}}

\newcommand{\etc}{\textit{etc}}

\newcommand{\Partial}[2]{\frac{\partial #1}{\partial #2}}

\DeclareMathOperator*{\argmin}{arg\,min}

\newcommand{\eqn}{Equation}
\newcommand{\Alg}{Algorithm}
\newcommand{\assum}{Assumption}
\newcommand{\lemm}{Lemma}
\newcommand{\theo}{Theorem}
\newcommand{\Prop}{Proposition}  
\newcommand{\Appendix}{Appendix}
\newcommand{\fig}{Figure}
\newcommand{\tbl}{Table}
\newcommand{\sect}{Section}

\newcommand{\alg}{\mathcal{A}lg^n}
\newcommand{\algest}{\mathcal{A}lg^{\tilde{n}}}

\newcommand{\prior}{\bm{\theta}}
\newcommand{\grad}{\bm{d}}
\newcommand{\param}{{\bm{\phi}}}               

\newcommand{\paramdiff}{\delta}
\newcommand{\momentum}{\gamma}

\newcommand{\nag}{Nesterov's accelerated gradient descent}
\newcommand{\nesg}{\gamma}

\newcommand{\nesgg}{\tilde{\gamma}}  
\newcommand{\neslr}{\frac{1}{\beta}}
\newcommand{\neslrn}[1]{\frac{#1}{\beta}}
\newcommand{\nesgradn}[1]{\nabla f(x_{#1})}
\newcommand{\nesgrad}{\nesgradn{t}}
\newcommand{\nesgradx}[1]{\nabla f(#1)}

\newcommand{\N}{\mathbb{N}}
\newcommand{\I}{\mathbb{I}}

\newcommand{\msi}{(k-1)n+1}  

\newcommand{\ms}[2]{#1\tiny{$\pm$#2}}
\newcommand{\0}{\hspace{0.47em}}
\newcommand{\loss}[1]{\mathcal{L}_{\tt{#1}}}
\newcommand{\lstot}{\loss{total}}
\newcommand{\lstfr}{\loss{tfr}}
\newcommand{\lsacc}{\loss{acc}}
\newcommand{\Data}[1]{\mathcal{D}_{\tt{#1}}}
\newcommand{\Dtr}{\Data{tr}}

\newcommand{\Dcub}{Bird}
\newcommand{\Dind}{Indoor}
\newcommand{\Dact}{Action}
\newcommand{\Ddog}{Dog}
\newcommand{\Domni}{Omniglot}
\newcommand{\Dmini}{\textit{mini}ImageNet}

\newcommand{\convex}{\alpha}  
\newcommand{\smooth}{\beta}  
\newtheorem{prop}{Proposition}

\newcommand{\paragrapht}[1]{\noindent\textbf{#1}}  
\usepackage{algorithm}
\usepackage{algorithmic}
\usepackage{caption}

\hypersetup{colorlinks,
citecolor=blue,
}

\def\beingshort
\usepackage[titletoc]{appendix}

\title{Multi-step Estimation for \\
Gradient-based Meta-learning}

\author{%
  Jin-Hwa Kim \\ 
  SK Telecom\\
  \texttt{jnhwkim@sk.com} \\
  \And
  Junyoung Park\\
  SK Telecom\\
  \texttt{jy1809.park@sk.com}
  \And
  Yongseok Choi\\
  SK Telecom\\
  \texttt{yongseokchoi@sk.com}
}

\begin{document}

\maketitle

\begin{abstract}
Gradient-based meta-learning approaches have been successful in few-shot learning, transfer learning, and a wide range of other domains. Despite its efficacy and simplicity, the burden of calculating the Hessian matrix with large memory footprints is the critical challenge in large-scale applications. To tackle this issue, we propose a simple yet straightforward method to reduce the cost by reusing the same gradient in a window of inner steps. We describe the dynamics of the multi-step estimation in the Lagrangian formalism and discuss how to reduce evaluating second-order derivatives estimating the dynamics. To validate our method, we experiment on meta-transfer learning and few-shot learning tasks for multiple settings. The experiment on meta-transfer emphasizes the applicability of training meta-networks, where other approximations are limited. For few-shot learning, we evaluate time and memory complexities compared with popular baselines. We show that our method significantly reduces training time and memory usage, maintaining competitive accuracies, or even outperforming in some cases.
\end{abstract}

\section{Introduction}
Meta-learning is a paradigm that improves the learning of knowledge for fast adaption to a novel task from learning tasks ~\cite{schmidhuber1987,hinton1987using,hochreiter2001learning,Andrychowicz2016}.
Meta-learning methods include model-based~\cite{Ravi2017miniimagenet,hochreiter2001learning,mishra2018a}, metric~\cite{vinyals2016matching,snell2017prototypical,sung2018learning,garcia2018few}, and optimization-based~\cite{Finn2017,Franceschi2018} approaches. Gradient-based method, one of optimization-based methods, exploits bi-level learning~\cite{Franceschi2017,Franceschi2018}, where inner-level tasks are solved by a gradient-based optimization for given knowledge (meta-parameters), while outer-level learns the common knowledge across multiple tasks (meta-objective).
Then, it performs error back-propagation~\cite{rumelhart1986} through the inner optimization path~\cite{lecun1988theoretical,Franceschi2017} for the meta-objective.
This is successfully applied to hyper-parameter optimization~\cite{Franceschi2017,Franceschi2018}, few-shot learning~\cite{Finn2017}, transfer learning~\cite{Jang2019}, \etc.

However, gradient-based methods face the challenge of evaluating second-order derivatives whose computational cost is proportional to the number of inner optimization steps.
This challenge is critical when the number of parameters or the meta-parameters forming networks (meta-networks)~\cite{Jang2019,Lorraine2020} is significantly increased for high-dimensional problems~\cite{domke2012,maclaurin2015,pedregosa2016,Franceschi2017,Franceschi2018}.
Due to the computational issues, meta-learning over long inner optimization steps is an active research area~\cite{Hospedales2020}.

In the paper, we provide a simple yet straightforward approximation to gradient-based meta-learning.
In \fig~\ref{fig:norm}, we observe that the difference between consecutive task-gradients asymptotically converges to zero in training.
Based on this, we take the first task-gradient in each non-overlapping window of inner-steps, and reuse the task-gradient in the inner optimization of the window to skip Hessian-vector products except one for each window, estimating the dynamics of inner optimization.
In \Prop~\ref{prop:multistep-short}, we show that our multi-step estimation is not compatible with a method merely decreasing the number of inner steps, and empirically validate the approximated dynamics retain the performance using multiple meta-transfer learning tasks and few-shot learning tasks.

For the related works, MAML approximation~\cite{Finn2017,Nichol2018,Raghu2020}, implicit gradient~\cite{Bengio2000,Rajeswaran2019,Lorraine2020}, and Hessian-free methods for meta-reinforcement learning (Meta-RL)~\cite{Song2020,Ji2020} can be considered.
The MAML approximations, First-order MAML~\cite{Finn2017}, Reptile~\cite{Nichol2018} including iMAML~\cite{Rajeswaran2019} find parameter initialization for fast adaptation sharing parameter space with meta-parameter's, which prohibits generalization to the other meta-learning approaches, \eg, learning meta-networks.
The implicit gradient methods decouple inner optimization trajectory with the calculation of meta-gradient. However, it is not suitable for large-scale applications since this requires the convergence of inner optimization (or its approximation) and induces the computational cost of matrix inversion.
The policy gradient in Meta-RL suffers a high variance from the sampling of trajectories for sparse rewards, so the Gaussian smoothing for hessian-free methods is favored in reinforcement learning~\cite{Song2020,Ji2020}. However, since Hessian-vector product can be efficiently performed using the reverse mode of automatic differentiation in practice~\cite{griewank1993some,Rajeswaran2019}, 
the sampling-based approaches less appealing to the other cases 
for not-too-high variance situation.

We summarize the contributions of this study as follows:
\begin{itemize}
    \item We approximate multi-step meta-learning by reusing the task-gradient in the consecutive inner-steps to avoid `full' Hessian-vector products, which is supported by empirical evidence that the normalized difference between consecutive task-gradients converges to zero.
    \item We efficiently reduce the training time for meta-transfer learning by 35\% and for few-shot learning by up to 50\%, while significantly reducing memory footprints, where the first-order and implicit-gradient methods cannot effectively apply to learn meta-networks.
    \item We validate our argument on the assorted benchmarks of meta-transfer learning tasks with six different transfer learning settings and few-shot learning tasks on \Domni~and \Dmini~datasets showing that competitive accuracies or even outperforming baselines.
\end{itemize}
\section{Preliminaries}

\subsection{Back-propagation through optimizing steps}

Consider the following bilevel optimization problem:
\begin{gather}
    \min_\theta ~\inf \{ \ls(s_{\theta}, \theta) | s_{\theta}\in \argmin_s E_{\theta} (s) \}
\end{gather}
where $\ls$ and $E_\theta$ are called {\it outer-objective} and {\it inner-objective}, respectively.
We view meta-learning as an instantiation of this framework~\cite{Franceschi2018}, and with this view $\ls$ is also called a {\it meta-objective}.
In a gradient-based approach~\cite{domke2012,maclaurin2015,Franceschi2017,Franceschi2018}, the inner problem is solved by a dynamical system illustrating gradient-based optimization.
Let the state (\eg, parameters $\phi$ and velocities $\vv$ for SGD with momentum) of optimization and the meta-parameter be $\vs_t$ and $\theta$, respectively. Then the dynamics $\Phi_t$ for a given $\theta$ is defined as follows~\cite{Franceschi2017}:
\begin{align}
    \vs_{t+1} &= \Phi_t (\vs_{t}; \theta)
\end{align}
for every $t \in \{1,...,T\}$. It describes a one-step optimization of the inner-objective $\ls_t(\phi_t;\theta)$ at $t$-step.
Notice that $\vs_1$ is related to $\theta$ for meta-objective (\eg, $\phi_1=\theta$ for MAML~\cite{Finn2017} or via inner-objective~\cite{Jang2019}).
Then, using the dynamics, the derivative of the meta-objective function $\ls$ with respective to the meta-parameter is:
\begin{align} \label{eq:meta-gradient}
    \Partial{\ls}{\theta} &= \sum_{t=1}^{T} 
        \Lambda_t^\intercal
        \Partial{\Phi_t(\vs_{t}; \theta)}{\theta}
\end{align}
where the Lagrangian multipliers $\Lambda_t$ (see \cite{lecun1988theoretical,Franceschi2017} for this interpretation) or back-propagated errors along with the steps are updated by
\begin{align} \label{eq:lagrangian}
    \Lambda_{t-1}^\intercal =
        \Lambda_{t}^\intercal \Partial{\Phi_t(\vs_t; \theta)}{\vs_t}, \hspace{1em}
     \Lambda_{T} = \Partial{\ls}{\vs_{T+1}}.
\end{align}
Refer \cite{Franceschi2017} for the derivation of this result. For example, but not restricted to, the stochastic gradient descent (SGD) with momentum is defined as:
\begin{align}
    \vv_{t+1}  \leftarrow \mu \vv_t + \vg_t + \omega\phi_t, ~~~
    \phi_{t+1} \leftarrow \phi_{t} - \eta \vv_{t+1}
\end{align}
where $\mu, \vg_t, \omega,$ and $\eta$ are momentum, $\nabla_\phi\ls_t(\phi_t;\theta)$, weight decay, and learning rate, respectively. Here, $\vs_t = [\phi_t, \vv_t]^\intercal$. The SGD instantiates the dynamics as follows:
\begin{align}
    \Partial{\ls}{\theta} &= \sum_{t=1}^{T} 
        \Lambda_t^\intercal
        \begin{bmatrix} 
            \Partial{\Phi_t(\vs_{t}, \theta)_\phi}{\theta} \\ 
            \Partial{\Phi_t(\vs_{t}, \theta)_\vv}{\theta} 
        \end{bmatrix}
        = \sum_{t=1}^T 
        \Lambda_t^\intercal
        \begin{bmatrix} -\eta \\ 1 \end{bmatrix} \Partial{\vg_t}{\theta} \label{eq_a1}
\end{align}
and the corresponding $\Lambda_t$ is updated by
\begin{align}
    \Lambda_{t-1}^\intercal &=
        \Lambda_{t}^\intercal
            \begin{bmatrix} 
                \Partial{\Phi_t(\vs_{t}; \theta)_\phi}{\phi_{t}} & \Partial{\Phi_t(\vs_{t}; \theta)_\phi}{\vv_{t}} \\ 
                \Partial{\Phi_t(\vs_{t}; \theta)_\vv}{\phi_{t}} & \Partial{\Phi_t(\vs_{t}; \theta)_\vv}{\vv_{t}}
            \end{bmatrix} =
        \Lambda_{t}^\intercal
            \begin{bmatrix} 
                1 - \eta\omega - \eta \Partial{\vg_t}{\phi_t} & -\eta\mu \\ 
                \omega + \Partial{\vg_t}{\phi_t} & \mu
            \end{bmatrix}. \label{eq_bc1} 
\end{align}
Here, we remark that the calculation of error back-propagation through optimizing steps is straightforward as previously discussed~\cite{lecun1988theoretical} since the optimizing steps are also differentiable.
Notice that $\partial{\vg}/\partial{\phi}$ is Hessian matrix and $\partial{\vg}/\partial{\theta}$ is second-order derivative with two variables, $\phi$ and $\theta$.
The computational bottleneck is to get the second-order derivatives; however, we can take advantage of Hessian-vector products using automatic differentiation~\cite{griewank1993some}, and it typically costs the five-fold computations and no more than twice memory of $\nabla\ls_t(\phi_t)$~\cite{Rajeswaran2019}.
For this reason, the sampling-based approximations of Hessian~\cite{Song2020,Ji2020} are prone to be costly in practice.

\subsection{Applications}

\paragrapht{Learning meta-networks.}
The meta-parameters of meta-networks usually do not share with the task-parameters.
However, the two groups of parameters can be inter-dependent through the loss functions in inner-loop, \ie, $\vg_t=\nabla_\phi\ls_t(\phi_t, \theta_k)$. The meta-update by meta-gradient is defined, using the previous equations, as follows:
\begin{align}
    \theta_{k+1} = \theta_{k} - \eta_{\text{meta}} \Partial{\ls(\phi_{T+1})}{\theta}.
\end{align}

\paragrapht{Few-shot learning.} The MAML algorithm~\cite{Finn2017} sets the initial task-parameters with the current meta-parameters as $\phi_1 \leftarrow \theta_k$ and the meta-parameters are updated by the meta-gradient as follows:
\begin{align}
    \theta_{k+1} = \theta_{k} - \eta_{\text{meta}} \frac{1}{M} \sum_{i=1}^M \Partial{\ls^{(i)}(\phi_{T+1}^{(i)})}{\theta}
\end{align}
where $\phi_{T+1}^{(i)}$ is the task-parameters after the $T$ steps of dynamics for the $i$-th task, and $M$ is the number of few-shot learning tasks. 
Since the MAML sets $\phi_1=\theta$ and the loss function of inner-loop is a function of $\phi$, we can rewrite \eqn~\ref{eq:meta-gradient} as follows:
\begin{align} \label{eq:meta-gradient-maml}
    \Partial{\ls}{\theta} &=
        \Lambda_1^\intercal
        \Partial{\Phi_{1}(\vs_{1})}{\vs_1}
        \Partial{\vs_1}{\theta}.
\end{align}

\section{Multi-step estimation}

\begin{wrapfigure}{r}{0.45\textwidth}
\vspace{-4em}
\begin{center}
    \includegraphics[width=0.45\textwidth]{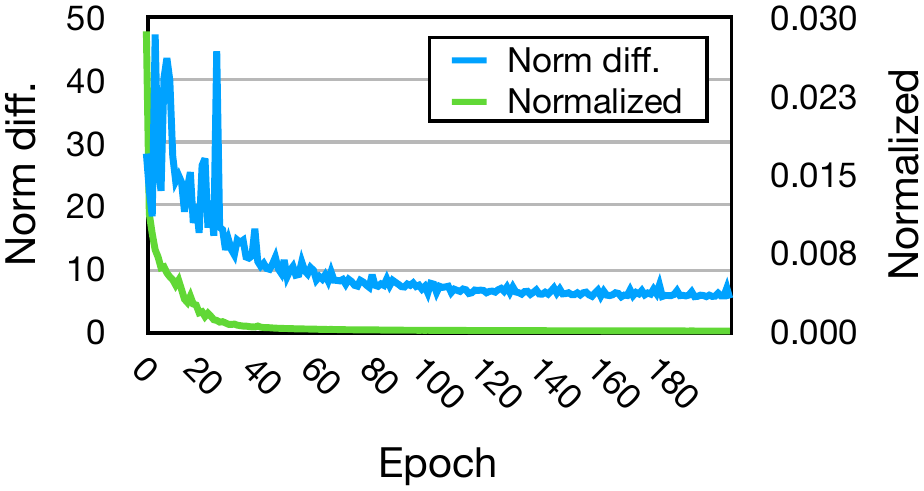}
\end{center}
\vspace{-1em}
\caption{Gradient difference in inner-step}
\label{fig:norm}
\end{wrapfigure}

\subsection{Motivation}
\label{sec:motivation}

Since the computational cost is proportional to the number of calculations of $\nabla^2_\phi\ls_t(\phi_t)$ and $\nabla_\theta\nabla_\phi\ls_t(\phi_t)$ in the inner-loop, we propose to approximate $\nabla\ls_{t'}(\phi_{t'}) \approx {\nabla}\ls_t(\phi_{t}) ~\forall t' \in \Delta_t^n = \{t, t+1, \cdots, t+n-1\}$, not over-wrapping the intervals, which means $\ls_{t' \in \Delta_t^n}$ are the same. In other words, we reuse $\nabla_\phi\ls_t(\phi_t)$ (\ie, $\vg_t$) in the dynamics.
The error of the approximation is minimized when $\| \nabla\ls_{t'}(\phi_{t'}) - \nabla\ls_{t''}(\phi_{t''}) \|$ is close to zero where $t'$ and $t''$ are in $\Delta_t^n$.
This assumption is validated in our experiment as shown in \fig~\ref{fig:norm}.
Where we measure the normalized norm of $\| \nabla\ls_t(\phi_{t+1}) - \nabla\ls_t(\phi_{t}) \|/\|\nabla\ls_t(\phi_{t+1}) \|$ in training, the difference diminishes to zero quickly in the early stage. The experimental details can be found in \Appendix~\ref{sec:norm}.
In the following section, we show that how this estimation shapes the dynamics. 

\subsection{Multi-step estimated dynamical system}  \label{sec:nstep-dynamics}

For the estimated system, we re-define the previous dynamical system as follows:
\begin{align}
    \label{eq:multi-step-estimated}
    \vs_{t+n} &= \hat{\Phi}_t^n(\vs_t;\theta)
\end{align}
where $\hat{\Phi}_t^n$ implicitly moves $n$ times with the fixed $\nabla_\phi\ls_t(\phi_t)$. Using the dynamics of the estimated $n$-step optimization, the meta-gradient in \eqn~\ref{eq:meta-gradient} and the Lagrangian multipliers are as follows:
\begin{align}
     \Partial{\ls}{\theta} \approx \sum_{k=1}^{K} 
        \Lambda_{kn}^\intercal
        \Partial{\hat{\Phi}_{\msi}^n(\vs_{\msi}; \theta)}{\theta}, \hspace{1em}
    \Lambda_{(k-1)n}^\intercal =
        \Lambda_{kn}^\intercal \Partial{\hat{\Phi}_{\msi}^n(\vs_{\msi}; \theta)}{\vs_{\msi}}
\end{align}
for $T = Kn$ where $K \in \N$.
It is noteworthy that the transformed dynamics effectively decrease the number of second-order derivative evaluations from $Kn$ to $K$. For the SGD with momentum,
\begin{align} \label{eq:mg-nsteps-sgdm}
    \Partial{\ls}{\theta} &\approx 
        \sum_{k=1}^{K} 
        \Lambda_{kn}^\intercal
        \bigg(
            \sum_{i=0}^{n-1}
            \begin{bmatrix} 
                1 - \eta\omega & -\eta\mu \\ 
                \omega         & \mu
            \end{bmatrix}^{i}
            \begin{bmatrix} 
                -\eta \\
                1
            \end{bmatrix}
            \Partial{\vg_{\msi}}{\theta}
        \bigg).
\end{align}
For the proof, please see \Appendix~\ref{sec:multi-step-sgdm}.
Letting $\omega = \mu = 0$, for the na\"ive SGD, we have that
\begin{align} \label{eq:mg-nsteps-sgd}
    \Partial{\ls}{\theta} &\approx 
        \sum_{k=1}^{K} 
        \lambda_{kn}^\intercal
        \bigg(
            -n\eta
            \Partial{\vg_{\msi}}{\theta}
        \bigg)
\end{align}
where $\Lambda$ is reduced to a vector $\lambda$ since $\mu=0$ as follows:
\begin{align} \label{eq:lambda-sgd}
    \lambda_{(k-1)n}^\intercal
    = 
    \lambda_{kn}^\intercal
    \big( 1 - n\eta \Partial{\vg_{\msi}}{\phi_{\msi}} \big), \hspace{1em}
    \lambda_{Kn}^\intercal = \Partial{\ls}{\phi_{Kn+1}}
\end{align}
therefore, for the na\"ive SGD, the multi-step estimated dynamical system is equivalent to the single-step system of two changed hyper-parameters $\eta' \leftarrow n\eta$ and $T' \leftarrow K = T/n$. However, be advised that, in general, the multi-step dynamics cannot substitute with the change of hyper-parameters due to the interaction of states as stated in Proposition~\ref{prop:multistep-short} and the proof can be found in \Appendix~\ref{sec:proof-multi-step}. An empirical validation in a meta-transfer learning task can be found in \Appendix~\ref{sec:supplementary_experiment_transfer}.

\begin{prop} \label{prop:multistep-short}
Multi-step estimation using SGD with momentum cannot be realized by one-step SGD with momentum consisting of different optimization hyper-parameters.
\end{prop}

A similar method~\cite{Goyal2017} is proposed for the classification task using a large mini-batch in a distributed environment. In this scenario, when the mini-batch size is multiplied by $n$ for the $n$ computing nodes, as a consequence, the number of iterations per epoch is decreased by $1/n$, they multiply the learning rate by $n$ to estimate the previous learning procedure. Note that we generalize for any differentiable optimization in a dynamical system, which can be of independent interest. The implementation of multi-step estimation is straightforward using the automatic differentiation of $\hat{\Phi}_t^n$ (\textit{ref.}~\Alg\ref{alg:meta-transfer}).
\section{Related work}
\label{sec:related_work}
In this section, we compare the approximation methods of meta-learning in terms of computational and memory complexities.
This comparison reveals the shortcomings of each method to apply for given meta-learning algorithms.

\paragrapht{First-order MAML~\cite{Finn2017}.} It assumes that the task-parameters $\phi$ are simply independent from the given meta-parameters $\theta$ in calculating the meta-gradient for each task in \eqn~\ref{eq:meta-gradient} as follows:
\begin{align} \label{eq:fo}
    \Partial{\ls}{\theta} &\approx 
        \Partial{\ls(\phi_{T+1})}{\phi}
\end{align}
where $\phi_{T+1}$ is the output of task-parameters via the $T$-step dynamics.
Notice that the meta-parameters could be in a different parameter space from the task-parameters' space for learning meta-networks, which inevitably deteriorates performance preventing to learn meta-networks.

\paragrapht{Reptile~\cite{Nichol2018}.} Similarly to the first-order MAML, the task-parameters are assumed to be independent from the meta-parameters. It ignores the gradients in the inner steps, rather it makes the meta-parameters slowly move toward the task-parameters as follows:
\begin{align} \label{eq:reptile}
    \Partial{\ls}{\theta} &\approx 
        \theta - \phi_{T+1}.
\end{align}
Note that Reptile and first-order MAML are identical under the proximal regularization~\cite{Rajeswaran2019}.

\paragrapht{iMAML~\cite{Rajeswaran2019}.}
The implicit MAML uses an implicit Jacobian exploiting a stationary inner solution $\phi_\star$ with respect to $\ls_t$ as follows:
\begin{align}
    \Partial{\ls}{\theta} &=
        \Partial{\phi_\star}{\theta}
        \nabla_\phi \ls(\phi_\star)
        =
        \big(\I + \frac{1}{\lambda}\nabla^2_\phi \ls_t(\phi_\star) \big)^{-1}
        \nabla_\phi \ls(\phi_\star)
\end{align}
where $\lambda$ is a hyper-parameter related to a proximal regularization for $\phi_\star$ to be close to $\theta$, and $\ls_t$ is a single loss function in the inner-loop. Since the Jacobian $\Partial{\phi_\star}{\theta}$ depends on $\nabla\ls_t(\phi_\star)$, this method decouples the meta-gradient computation with inner optimization trajectory.
The limitations are additional computations to find a inner-level solution and approximating an inversion of matrix, which are unfavorable for a large-scale application.

\paragrapht{Other Hessian-free methods.}
HF-MAML~\cite{Fallah2019} exploits the first-order approximation of Hessian-vector product~\cite{Martens2010} for one-step MAML. More recently, ES-MAML~\cite{Song2020} uses a zero-order smoothing method for the Hessian approximation, while it has a large estimation error with a small number of samplings. GGS-MAML~\cite{Ji2020} proposes a gradient-based Gaussian smoothing (GGS) method as a variant. 
The MAML-based approaches including the iMAML cannot approximate the second-order derivatives with two variables, meta and task-parameters, $\nabla_\theta\nabla_\phi\ls(\phi)$, which are used to learn meta-networks.
And, as mentioned before, since automatic differentiation efficiently computes Hessian-vector products in practice~\cite{Rajeswaran2019}, these Hessian-free methods are less beneficial to our view.
\section{Experiments}

\subsection{Meta-transfer learning}

\begin{algorithm}[tb]
   \caption{Multi-step estimated meta-transfer learning (\textit{cf.} Algorithm 1 of the original work~\cite{Jang2019})}
   \label{alg:meta-transfer}
\begin{algorithmic}[1]
   \STATE {\bfseries Input:} Training examples $\Dtr=\{(x_i,y_i)\}$
   \STATE {\bfseries Hyper-parameters:} Mini-batch size $B$, optimizer state $v$, inner steps $T=KN$
   \REPEAT
   \STATE Sample a batch $\mathcal{B}\subset\mathcal{D}_{\tt train}$ with $|\mathcal{B}|=B$
   \STATE Update $\phi$ to minimize $\frac{1}{B}\sum_{(x,y)\in\mathcal{B}}\lstot(\phi|x,y,\theta)$
   \STATE Initialize $s_0 \gets (\phi, v)$ to begin inner-steps
   \FOR{$k=0$ {\bfseries to} $K-1$}
    \STATE $g_{Nk} \gets \nabla_\phi \lstfr(\phi_{Nk}; \theta, x \in \mathcal{B})$
   \FOR{$n=0$ {\bfseries to} $N-1$}
   \STATE $s_{Nk+n+1} \gets \Phi_{Nk+n}\big( s_{Nk+n}; \theta, g_{Nk} \big)$
   \smash{\raisebox{0.15\dimexpr4\baselineskip+4\itemsep+2\parskip}{\hspace{.2em}$\left.\rule{0pt}{.5\dimexpr4\baselineskip+4\itemsep+2\parskip}\right\}$\hspace{.2em} \parbox{6.5cm}{
   This part describes $ \hat{\Phi}_{Nk}^N$ in Eqn.~\ref{eq:multi-step-estimated} \\ which estimates $s_{N(k+1)}$.
   }}}
   \ENDFOR
   \ENDFOR
   \STATE $s_{T+1} \gets \Phi_T\big( s_{T}; \theta, \nabla_\phi \lsacc(\phi_T; (x,y) \in \mathcal{B}) \big)$
   \STATE Meta-update $\theta$ using $\nabla_\theta\frac{1}{B}\sum_{(x,y)\in\mathcal{B}}\lsacc(\phi_{T+1}|x,y)$ through $\phi_t$
   \UNTIL{done}
\end{algorithmic}
\end{algorithm}

\paragrapht{Task.} To validate the efficacy of our method, we apply our method to a recently proposed meta-learning to learn meta-networks $\theta$ which control the transferability to task-networks $\phi$ in transfer learning~\cite{Jang2019}. 
It transfers the acquired knowledge of a source model from training with a large source dataset to the other task having possibly different target model architecture and a relatively smaller target dataset, using the following compound loss with a weight of $\beta$ (we set $\beta=0.5$):
\begin{align} \label{eq:meta-transfer}
    \loss{total}(\phi|\Dtr,\theta) &= \lsacc(\phi|\Dtr) + \beta \lstfr(\phi|\Dtr,\theta)
\end{align}
where $\lsacc$ and $\lstfr$ denote a classification loss and a transfer loss, respectively. $\Dtr$ is the examples from the training split for each corresponding dataset.

\paragrapht{Meta-networks.} The meta-networks involve in $\lstfr$ through $\theta$.
For each pair of intermediate representations of source and target models for the same target input, $\lstfr$ measures a transfer loss, weighting with nesting two groups of learnable parameters in meta-networks for pair-wise (where) and channel-wise (what) transferability. Note that since it transfers the knowledge across heterogeneous architectures, small networks are used to match between the dimensions of representations from source and target models.
For ResNets, we take the outputs of each stage (three stages of ResNet-32 and four of ResNet-34), while, for VGG-9, the inputs of the five down-scaling layers.
Please refer to their work~\cite{Jang2019} and the code~\footnote{\url{https://github.com/alinlab/L2T-ww}}. Unless stated otherwise, we follow their experimental setting, including how to set aside validation splits (\ie, 10\% of training data) to select model.

\paragrapht{Meta-algorithm.} The \Alg~\ref{alg:meta-transfer} includes $T=KN$ consecutive transfer losses, which is the point we estimate.
Notice that, in the original paper~\cite{Jang2019}, they formally validate the efficacy of meta networks and the proposed bi-level scheme with two separated inner-objective functions (Line 10 and 13 in \Alg~\ref{alg:meta-transfer}) in their Appendix C.1 and C.2 showing significant improvements in multiple tasks.
They empirically find that $T=2$ is optimal (we also observe it in \Appendix~\ref{sec:supplementary_experiment_transfer}), and we estimate with $K = 1$.
We emphasize that the transfer losses are for the meta-networks controlling transferability, the feature matching networks, and target models, which are used to calculate the meta-gradient.

\begin{wraptable}{r}{0.45\textwidth}
  \vspace{-1em}
  \centering
  \caption{Time elapse for one epoch (second) using a Titan Xp. $\pm$ denotes the standard deviation of three models for \Dcub~\cite{wah2011caltech}.
  Eventually, our method reduces 7+ hours (35\%) for 200 epochs with competitive accuracy.
  }\label{tbl:time-transfer}
  \vspace{0.1in}
  \begin{tabular}{cccc}
  \toprule
  Method & Elapse & Total & Ratio \\ 
  \midrule
  Baseline~\cite{Jang2019} & \ms{379.7}{30.9}       & 21h & 1.00 \\
  Multi-step & \bf{\ms{248.7}{14.2}}  & \bf{14h} & \bf{0.65} \\
  \bottomrule
  \end{tabular}
\end{wraptable}

\paragrapht{Domains and models.} 
Following the previous~\cite{Jang2019}, we validate on six datasets, where two datasets have the small-sized inputs of 32x32, the others have the large-sized inputs of 224x224.
For the small-sized inputs, TinyImageNet~\cite{tinyimagenet} is used as a source domain, while CIFAR-100~\cite{krizhevsky2009tiny} and STL-10~\cite{coates2011analysis} as target domains. ResNet-32~\cite{He2015,Jang2019} and VGG-9~\cite{Simonyan2015,Srinivas2018} are the source and target models, respectively.
For the second part using the 224x224 inputs,
ImageNet~\cite{deng2009imagenet} is used as a source domain, while Caltech-UCSD Bird 200 (\Dcub)~\cite{wah2011caltech}, MIT Indoor Scene Recognition (\Dind)~\cite{quattoni2009recognizing}, Stanford 40 Actions (\Dact)~\cite{yao2011human} and Stanford Dogs (\Ddog)~\cite{Khosla2011dog} as target domains. ResNet-34 and ResNet-18~\cite{He2015} are the source and target models, respectively.
We resize and crop images if necessary (\Appendix~\ref{sec:preproc}).

\paragrapht{Time complexity.}
\tbl~\ref{tbl:time-transfer} shows that our method significantly reduces training time over 7 hours (35\%), and the same tendency can be found across multiple benchmarks (\Appendix~\ref{sec:supplementary_transfer_time}).
Although meta-learning greatly boosts the performance of transfer learning~\cite{Jang2019}, the increased training time was one of drawbacks of meta-transfer learning.
Thus from this result, we argue that our method is a valuable approach in practice. For the results with $T=1,3$, please refer to \fig~\ref{fig:cub200_comp} in \Appendix~\ref{sec:supplementary_experiment_transfer}.

\paragrapht{Accuracy.}
To assess our multi-step estimation, along with momentum SGD in \tbl~\ref{tbl:sgd}, we use Adam~\cite{Kingma2014} for its adaptive gradient dynamics in \tbl~\ref{tbl:adam} (\Appendix~\ref{sec:hyper-transfer} for hyper-parameters).
\textit{Multi-step} estimation (ours) gives competitive results compared with \textit{Baseline}~\cite{Jang2019} on both optimizers through extensive benchmarks. 
We speculate that, empirically, our assumption in \sect~\ref{sec:motivation} on the convergence of gradient difference does not cause any significant performance drop, and our multi-step dynamical system possibly works well with the optimizers other than SGD, \ie, Adam.

\begin{table*}[t!]
  \centering
  \caption{Classification accuracy (\%) of the transferred models using the SGD with momentum (inner-level).
  \textit{Scratch} does not perform transfer learning.
  \textit{Baseline}~\cite{Jang2019} is L2T-ww (all-to-all) in their paper.
   $\pm$ denotes the standard deviation of three models.
  }\label{tbl:sgd}
  \vspace{0.1in}
  \begin{tabular}{ccccccc}
  \toprule
  Source &  \multicolumn{2}{c}{TinyImageNet} & \multicolumn{4}{c}{ImageNet}      \\ 
  \cmidrule(lr){1-1} \cmidrule(lr){2-3} \cmidrule(lr){4-7}
  Target  &  CIFAR-100  & STL-10 & \Dcub & \Dind & \Dact & \Ddog \\ \midrule
  Scratch                  & \ms{67.69}{0.22}      & \ms{65.18}{0.91}      & \ms{42.15}{0.75}      & \ms{48.91}{0.53}      & \ms{36.93}{0.68}      & \ms{58.08}{0.26}  \\
  Baseline~\cite{Jang2019} & \ms{70.96}{0.61}      & \bf{\ms{78.31}{0.21}}      & \bf{\ms{65.05}{1.19}} & \ms{64.85}{2.75}      & \bf{\ms{63.08}{0.88}}      & \ms{78.08}{0.96}  \\
  Multi-step               & \bf{\ms{70.97}{0.38}}      & \ms{77.83}{0.74}      & \ms{64.89}{0.98}      & \bf{\ms{66.67}{0.56}} & \ms{61.93}{3.32}      & \bf{\ms{78.15}{0.22}} \\
  \bottomrule
  \end{tabular}
\end{table*}
\begin{table*}[t!]
  \centering
  \caption{Classification accuracy (\%) of the transferred models using the Adam~\cite{Kingma2014} (inner-level).
  }\label{tbl:adam}
  \vspace{0.1in}
  \begin{tabular}{ccccccc}
  \toprule
  Source &  \multicolumn{2}{c}{TinyImageNet} & \multicolumn{4}{c}{ImageNet}      \\ 
  \cmidrule(lr){1-1} \cmidrule(lr){2-3} \cmidrule(lr){4-7}
  Target &  CIFAR-100  & STL-10 & \Dcub & \Dind & \Dact & \Ddog \\ \midrule
  Scratch                  & \ms{67.69}{0.22}      & \ms{65.18}{0.91}      & \ms{42.15}{0.75}      & \ms{48.91}{0.53}      & \ms{36.93}{0.68}      & \ms{58.08}{0.26}  \\
  Baseline (ours)          & \ms{69.35}{0.09}      & \bf{\ms{80.61}{0.29}}      & \ms{66.26}{0.51}      & \bf{\ms{67.74}{0.64}} & \ms{62.05}{2.15}      & \ms{74.38}{0.45}  \\
  Multi-step               & \bf{\ms{70.12}{0.23}}      & \ms{79.58}{0.54}      & \bf{\ms{66.51}{1.08}} & \ms{67.14}{1.08}      & \bf{\ms{65.43}{0.23}} & \bf{\ms{77.65}{0.11}} \\
  \bottomrule
  \end{tabular}
\end{table*}

\paragrapht{Other approximations.}
First-order MAML and Reptile cannot be applied to \Alg~\ref{alg:meta-transfer} due to their assumption for parameter-initialization approach.
Yet, an implementation of First-order MAML where the meta-objective is replaced with $\loss{total}$ instead of $\lsacc$ to get $\nabla_\theta \loss{total}(\phi_{T+1})\neq0$ (note that $\nabla_\theta \lsacc(\phi_{T+1})=0$) 
can be considered.
Notice that the gradient is not computed through $\phi_{t<T+1}$ for first-order approximation.
This method severely underperforms (41.46$\pm$2.34) compared with 
\textit{Multi-step} estimation (64.89$\pm$0.98) for \Dcub~experiment.
The iMAML cannot apply due to the assumption of costly inner-convergence, and sampling-based methods for Hessian are less beneficial by the trade-off between the number of samplings, related to accuracy, and computational cost.

\subsection{Few-shot learning}

\begin{table*}[t!]
  \centering
  \caption{\Domni~few-shot classification accuracy (\%).
  The second section uses our implementation for a fair comparison.
  Multi-step ($N$) uses the same second order derivatives for $N$ consecutive inner updates.
  $\dagger$ denotes the meta-gradient is computed using Hessian-free method and proximal regularization~\cite{Rajeswaran2019}, but the others do not apply this regularization.
  $\pm$ denotes 95\% confidence interval. 
  }\label{tbl:omniglot}
  \vspace{0.1in}
  \begin{tabular}{ccccccc}
  \toprule
  Method &  5-way 1-shot & 5-way 5-shot & 20-way 1-shot & 20-way 5-shot      \\ \midrule
  MAML~\cite{Finn2017}      & \ms{98.7\0}{0.4\0}   & \bf{\ms{99.9\0}{0.1\0}} & \ms{95.8\0}{0.3\0}    & \ms{98.9\0}{0.2\0}  \\
  FO-MAML~\cite{Finn2017}   & \ms{98.3\0}{0.5\0}   & \ms{99.2\0}{0.2\0}      & \ms{89.4\0}{0.5\0}    & \ms{97.9\0}{0.1\0}  \\
  Reptile~\cite{Nichol2018} & \ms{97.68}{0.04} & \ms{99.48}{0.06}    & \ms{89.43}{0.14}  & \ms{97.12}{0.32}  \\ 
  iMAML$^\dagger$~\cite{Rajeswaran2019} & \bf{\ms{99.50}{0.26}} & \ms{99.74}{0.11} & \ms{96.18}{0.36} & \bf{\ms{99.14}{0.10}}  \\ 
  \midrule 
  MAML (ours)           & \bf{\ms{99.10}{0.77}} & \bf{\ms{99.66}{0.44}} & \bf{\ms{95.80}{0.81}} & \bf{\ms{98.73}{0.37}}  \\
  FO-MAML (ours)        & \ms{98.40}{1.11}       &       \ms{99.46}{0.58}    &   \ms{90.53}{1.13}    &       \ms{96.73}{0.55}    \\
  Multi-step (2)        & \ms{99.10}{0.81}      & \ms{99.66}{0.43}       & \ms{95.60}{0.78}      & \ms{98.70}{0.38}  \\
  Multi-step (4)        & \ms{98.97}{1.02}     & \ms{99.66}{0.47}       & \ms{95.75}{0.77}     & \ms{98.60}{1.03}  \\
  Multi-step (8)        & \ms{98.70}{1.03}      & \ms{99.50}{0.52}        & \ms{94.24}{0.95}     & \ms{97.70}{0.51}  \\
  \bottomrule
  \end{tabular}
\end{table*}

\paragrapht{Task.}
To further validate our method, we study on few-shot learning task, following a standard N-way K-shot protocol~\cite{Finn2017}.
We use \Domni~\cite{Lake2011omniglot} and \Dmini~\cite{Ravi2017miniimagenet} datasets, which are popular in the literatures.
To focus on the analysis of computational complexity (time and memory),
we compare with MAML~\cite{Finn2017}, FO-MAML (First-order MAML)~\cite{Finn2017}, Reptile~\cite{Nichol2018} and iMAML~\cite{Rajeswaran2019}
, since these variants of MAML methods are well-studied for the complexity comparison.

\paragrapht{Experimental details.}
We follow the previous experimental setup~\cite{Finn2017} in terms of neural network architecture and data preprocessing.
We first tried reproducing MAML~\cite{Finn2017} following all the same settings described in the paper. However, we could not have the numbers, especially for the 20-way 1-shot task on \Domni. Notice that this issue is also reported in~\cite{antoniou2018how, Abdullah2018}.
Instead, we use the SGD with momentum optimizer for the inner updates in lieu of vanilla SGD~\cite{Finn2017} to match with the scores in the paper.
We denote Multi-step ($N$) as our multi-step estimation method reusing $\nabla_\phi\ls_t(\phi_t)$ $N$-times for the inner updates.
To apply our method, we take eight inner updates and compare Multi-step (2), Multi-step (4) and Multi-step (8).
The average accuracy and 95\% confidence interval (CI) are computed by averaging on 600 test episodes.
Please refer \Appendix~\ref{sec:hyper-few-shot} for further experimental details.

\paragrapht{\Domni~performance.}
\tbl~\ref{tbl:omniglot} shows the performance of the baselines and our methods on \Domni~dataset.
Ours significantly outperforms first-order methods like FO-MAML and Reptile, especially on 20-way 1-shot (more than 5\% accuracy gap), and competitive with MAML and iMAML on all cases.
The performances of iMAML are slightly better than ours, but that may due to longer inner updates (iMAML uses 16 and 25 update steps for 5-way, 20-way experiments, respectively) or the usage of proximal regularization~\cite{Rajeswaran2019}, or both.
Note that the performances of Multi-steps ($N$) are slightly decreasing as $N$ is increasing as expected, but they retain their accuracies compared with MAML (ours).
In most cases, the margin of performance is under 0.4\% except Multi-step (8) on 20-way tasks, which are at most 1.56\%.

\begin{figure}[!t]
\begin{minipage}{.48\textwidth}
    \small
    \vspace{-1em}
    \begin{center}
        \includegraphics[width=\textwidth]{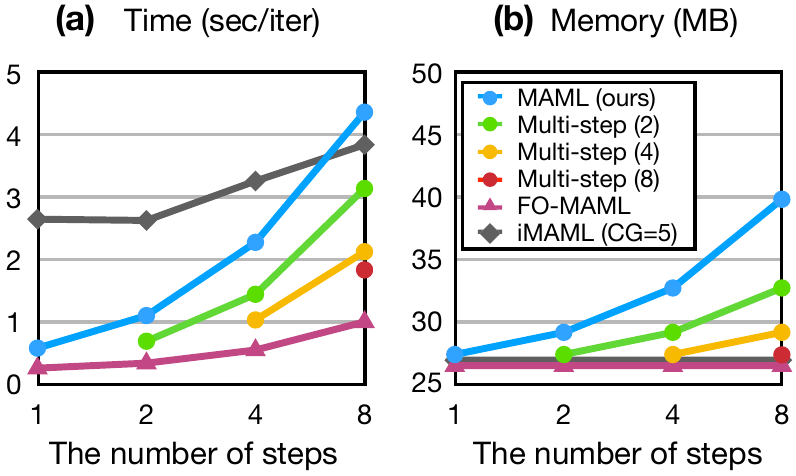}
    \end{center}
    \vspace{-.5em}
    \captionof{figure}{Memory usage and training time of 5-way 1-shot task for \Domni. 
        Best viewed in color.
        For a fair comparison, our code is based on the MAML implementation from \url{https://github.com/dragen1860/MAML-Pytorch}.}
    \label{fig:mem_speed}
\end{minipage}
\hspace{1em}
\begin{minipage}{.48\textwidth}
    {\small
  \captionof{table}{Few-shot classification accuracy (\%) for \Dmini.
  The second section uses our implementations for a fair comparison.
  $\dagger$ uses Hessian-free method and proximal regularization~\cite{Rajeswaran2019}.
  $\pm$ denotes 95\% confidence interval.
  }\label{tbl:miniimagenet}
  \vspace{-0.3em}
  \begin{center}
  \begin{tabular}{ccccc}
  \toprule
  Method &  5-way 1-shot & 5-way 5-shot       \\ \midrule
  MAML~\cite{Finn2017}      & \ms{48.70}{1.84}   & \ms{63.11}{0.92}  \\
  FO-MAML~\cite{Finn2017}   & \ms{48.07}{1.75}   & \ms{63.15}{0.91}  \\
  Reptile~\cite{Nichol2018} & \bf{\ms{49.97}{0.32}} & \bf{\ms{65.99}{0.58}} \\ 
  iMAML$^\dagger$~\cite{Rajeswaran2019} & \ms{49.30}{1.88} & -   \\ 
  \midrule 
  MAML (ours)           & \ms{48.49}{0.83}      & \ms{63.43}{0.71}   \\
  FOMAML (ours)         & \ms{44.90}{0.75}      &               \ms{61.57}{0.68}    \\
  Multi-step (2)        & \ms{48.68}{0.82}      & \bf{\ms{63.70}{0.73}} \\
  Multi-step (4)        & \bf{\ms{48.90}{0.82}} & \ms{63.67}{0.75}   \\
  Multi-step (8)        & \ms{46.85}{0.79}      & \ms{60.84}{0.71}   \\
  \bottomrule
  \end{tabular}
  \end{center}
    }
\end{minipage}
\end{figure}

\paragrapht{\Dmini~performance.}
\tbl~\ref{tbl:miniimagenet} shows the result on \Dmini.
Similarly, Multi-steps ($N$) retain performances when $N\le4$, but not for $N=8$ (2.59\% mean accuracy drop on 5-way 5-shot task), which shows a limit of our method depending on the number of estimation steps.
Reptile is the best among all methods on \Dmini, but, interestingly, the approximation method already surpasses MAML.
The gain may attribute to the change of inner optimizer to Adam~\cite{Kingma2014} and other hyper-parameters.
Finally, we emphasize that our method smoothly controls the amount of approximation using $N$, outperforming or being competitive with other baselines.

\begin{figure*}
    \begin{center}
        \includegraphics[width=.9\textwidth]{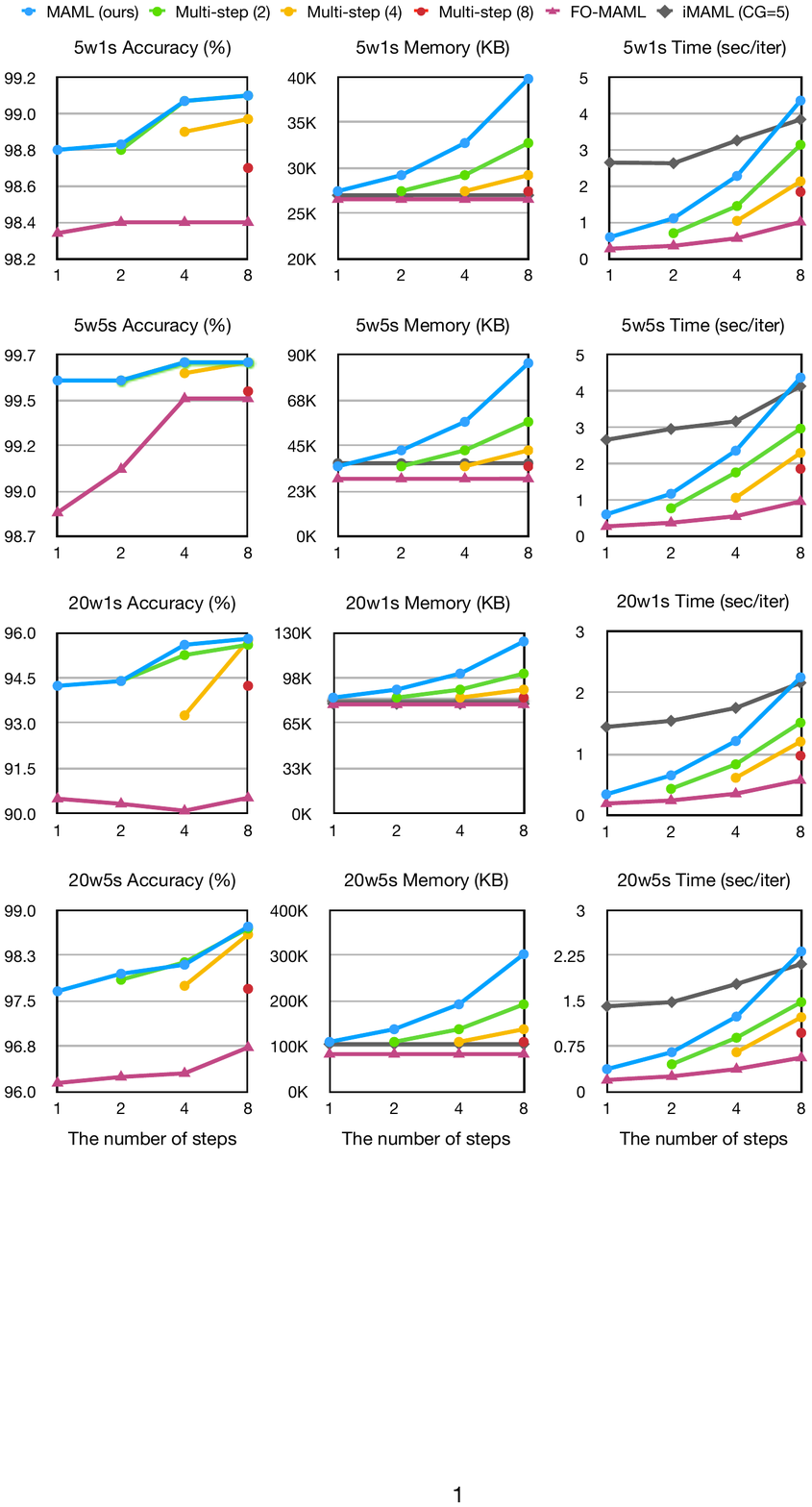}
    \end{center}
    \captionof{figure}{Memory usage and training time of the few-shot tasks for \Domni. 
        For a fair comparison, we use the MAML implementation code from \url{https://github.com/dragen1860/MAML-Pytorch}. Please refer to \tbl~\ref{tbl:omniglot} for the accuracy of iMAML~\cite{Rajeswaran2019}.
        }
    \label{fig:mem_speed_all}
\end{figure*}

\paragrapht{Time and memory.}
We measure time and memory usage of various algorithms for 5-way 1-shot on \Domni, as shown in \fig~\ref{fig:mem_speed} (for more results, \fig~\ref{fig:mem_speed_all} in \Appendix).
Note that Multi-steps ($N$) significantly reduces memory and time consumption as $N$ increases.
In particular, compared with MAML with eight update steps, Multi-steps (4) is two times faster and uses significantly less memory with just 0.1\% accuracy drop.
FO-MAML is an efficient method in both time and space complexities since it does not need to save the intermediate states of inner steps and to calculate second-order derivatives through updating trajectories.
However, it is likely to underperform compared to the others, as shown in \tbl~\ref{tbl:omniglot}.
For the same reason, iMAML uses less memory as much as first-order methods; however, it suffers the increased time in our experiments of less-than-eight inner-steps due to iterative computations for inverting a matrix to estimate the meta-gradients for each task.


\section{Conclusions}
We propose a simple yet robust method to estimate multi-step meta-learning, lazily updating the gradients of inner-steps to minimize the cost of time and memory significantly. 
Compared with the other approximations, our method is more general since it does not assume the share of meta and task-parameter spaces as in parameter initialization approaches, and does not rely on iterative methods to approximate meta-gradients or Hessian matrix which have a computational downside.
For the challenging meta-transfer learning tasks on several datasets, our method significantly reduces meta-training time while maintaining competitive accuracies or even outperforming in some configurations.
For the few-shot learning tasks, we deeply analyze on time and memory complexities to highlight on the gradual approximation with the hyper-parameter of $N$, reassuring our results on meta-transfer learning.
However, notice that our method may not work well under the circumstance where task-parameters are re-initialized irrespective of meta-parameters for each task optimization~\cite{Franceschi2018}, since, in that case, the assumption observed in \fig~\ref{fig:norm} may not hold.

\section*{Broader Impact}

This work proposes an efficient meta-learning method retaining competitive performances which potentially saves computing resources for our environment. We believe the benefit or disadvantage from this research is not particularly limited to a certain group.
Our method does not leverage the biases in the data validating on various meta-transfer learning settings and the few-shot learning experiments on \Domni~and \Dmini.



\subsubsection*{Acknowledgments}

The authors would like to thank Dong-Yeon Cho for helpful feedbacks to polish Abstract and Introduction.

\subsubsection*{Author Contributions}

J.-H.K. initiated the work as the corresponding author, proposed the methods, performed the experiments, and wrote and edited the draft of paper.
J.P. contributed to Lemma 1, Proposition 1, Section 5.2, performed the experiments, and helped editing the paper.
Y.C. performed the experiments and contributed to editing the paper.

\bibliography{main}
\bibliographystyle{unsrtnat}

\newpage
\setcounter{section}{0}
\renewcommand\thesection{\Alph{section}}
\renewcommand\thesubsection{\thesection.\arabic{subsection}}

\section{The gradient difference in the inner-step}
\label{sec:norm}
For the measurement of the gradient difference between two steps in the inner-loop, we perform the transfer learning using meta-learning~\cite{Jang2019}. The task is to transfer the knowledge of ResNet-34 pretrained on the ImageNet~\cite{russakovsky2014imagenet} to ResNet-18 for the CUB200 dataset~\cite{wah2011caltech}. We follow the experimental protocol in the paper. We measure the two-norm of the gradient difference for the matching loss in the inner-step.

\section{Multi-step estimation of SGD with momentum}
\label{sec:multi-step-sgdm}

In \sect~\ref{sec:nstep-dynamics}, the dynamics of the estimated $n$-step optimization define the meta-gradient as follows:
\begin{align}
     \Partial{\ls}{\theta} \approx \sum_{k=1}^{K} 
        \Lambda_{kn}^\intercal
        \Partial{\hat{\Phi}_{\msi}^n}{\theta}, \hspace{1em}
    \Lambda_{(k-1)n}^\intercal =
        \Lambda_{kn}^\intercal \Partial{\hat{\Phi}_{\msi}^n}{\vs_{\msi}}
\end{align}
where $\hat{\Phi}_{\msi}^n:=\hat{\Phi}_{\msi}^n(\vs_{\msi}; \theta)$.
For the SGD with momentum,
\begin{align} \label{eq:mg-nsteps-sgdm-appendix}
    \Partial{\ls}{\theta} &\approx 
        \sum_{k=1}^{K} 
        \Lambda_{kn}^\intercal
        \bigg(
            \begin{bmatrix} 
                1 - \eta\omega & -\eta\mu \\ 
                \omega         & \mu
            \end{bmatrix}
            \Partial{\hat{\Phi}_{\msi}^{n-1}}{\theta}
            + 
            \begin{bmatrix} 
                -\eta \\
                1
            \end{bmatrix}
            \Partial{\vg_{\msi}}{\theta}
        \bigg) \\
    &= \sum_{k=1}^{K} 
        \Lambda_{kn}^\intercal
        \bigg(
            \begin{bmatrix} 
                1 - \eta\omega & -\eta\mu \\ 
                \omega         & \mu
            \end{bmatrix}^{n-1}
            \Partial{\Phi_{\msi}}{\theta}
            + 
            \sum_{i=0}^{n-2}
            \begin{bmatrix} 
                1 - \eta\omega & -\eta\mu \\ 
                \omega         & \mu
            \end{bmatrix}^{i}
            \begin{bmatrix} 
                -\eta \\
                1
            \end{bmatrix}
            \Partial{\vg_{\msi}}{\theta}
        \bigg) \\
    &= \sum_{k=1}^{K} 
        \Lambda_{kn}^\intercal
        \bigg(
            \sum_{i=0}^{n-1}
            \begin{bmatrix} 
                1 - \eta\omega & -\eta\mu \\ 
                \omega         & \mu
            \end{bmatrix}^{i}
            \begin{bmatrix} 
                -\eta \\
                1
            \end{bmatrix}
            \Partial{\vg_{\msi}}{\theta}
        \bigg)
\end{align}
where the first equation uses chain rule for the one estimated step, while the last terms in the equations in the parentheses are for the reused $\vg_{\msi}$ in the $n-1$ steps. For the na\"ive SGD, \ie, $\omega = \mu = 0$,
\begin{align} \label{eq:mg-nsteps-sgd-appendix}
    &=
    \sum_{k=1}^{K} 
        \Lambda_{kn}^\intercal
        \bigg(
            \begin{bmatrix} 
                1 & 0 \\ 
                0 & 0
            \end{bmatrix}^{n-1}
            \Partial{\Phi_{\msi}}{\theta}
            + 
            \sum_{i=0}^{n-2}
            \begin{bmatrix} 
                1 & 0 \\ 
                0 & 0
            \end{bmatrix}^{i}
            \begin{bmatrix} 
                -\eta \\
                1
            \end{bmatrix}
            \Partial{\vg_{\msi}}{\theta}
        \bigg) \\
    &=
    \sum_{k=1}^{K} 
        \Lambda_{kn}^\intercal
        \bigg(
            \begin{bmatrix} 
                -\eta \\
                0
            \end{bmatrix}
            \Partial{\vg_{\msi}}{\theta}
            + 
            \begin{bmatrix} 
                -(n-1)\eta \\
                1
            \end{bmatrix}
            \Partial{\vg_{\msi}}{\theta}
        \bigg) \\
    &=
        \sum_{k=1}^{K} 
        \lambda_{kn}^\intercal
        \bigg(
            -n\eta
            \Partial{\vg_{\msi}}{\theta}
        \bigg)
\end{align}
where the last equation comes from the definition of $\Lambda_T$ and $\lambda$ is defined as follows:
\begin{align}
    \lambda_{(k-1)n}^\intercal
    = 
    \lambda_{kn}^\intercal
    \big( 1 - n\eta \Partial{\vg_{\msi}}{\phi_{\msi}} \big), \hspace{1em}
    \lambda_{Kn}^\intercal = \Partial{\ls}{\phi_{Kn+1}}.
\end{align}

\section{Proof that multi-step estimation is not one-step with different hyper-parameters}
\label{sec:proof-multi-step}
Recall the formulas defining SGD with momentum:
\begin{align}
    \vv_{t+1}  &= \mu \vv_t + \vg_t + \omega\phi_t \label{sgdmdef-v}\\
    \phi_{t+1} &= \phi_{t} - \eta \vv_{t+1} \label{sgdmdef-phi}
\end{align}
with $\vv_1=0$, $\phi_1=\phi$ is given, and $\vg_t=\nabla_{\phi_t}\ls_t(\phi_t)$.

\begin{lemma} \label{lem:SGD-recur}
Assuming $\vg_t=\vg$ for all $t\ge 1$, $\vv_t$ and $\phi_t$ can be represented as linear combinations of $\vg$ and $\phi$. In other words, we may write
\begin{align}
    \vv_t = b_t^\vv \vg + c_t^\vv \phi \\
    \phi_t = b_t^\phi \vg + c_t^\phi \phi
\end{align}
where $b_t^\vv, c_t^\vv, b_t^\phi$ and $c_t^\phi$ can be written in terms of $\mu, \omega$ and $\eta$.
Furthermore, we have the following recurrence relations:
\begin{align}
    b_{t+1}^\vv  &= \mu b_t^\vv + 1 + \omega b_t^\phi \\
    c_{t+1}^\vv  &= \mu c_t^\vv + \omega c_t^\phi \\
    b_{t+1}^\phi &= -\eta \mu b_t^\vv - \eta + (1-\eta\omega)b_t^\phi \\
    c_{t+1}^\phi &= -\eta \mu c_t^\vv + (1-\eta\omega)c_t^\phi
\end{align}
for $t\ge 1$.
\end{lemma}
\begin{proof}
By induction using~(\ref{sgdmdef-v}) and (\ref{sgdmdef-phi}), the first assertion is clear. Note that
\begin{align}
    b_{t+1}^\vv\vg + c_{t+1}^\vv\phi &= \vv_{t+1} \\
                                    &= \mu \vv_t + \vg + \omega \phi_t \\
                                    &= \mu (b_t^\vv \vg + c_t^\vv \phi) + \vg + \omega (b_t^\phi \vg + c_t^\phi \phi) \\
                                    &= (\mu b_t^\vv + 1 + \omega b_t^\phi)\vg + (\mu c_t^\vv + \omega c_t^\phi) \phi.
\end{align}
Since this should hold for any $\vg$ and $\phi$, their coefficients should coincide.
Similarly,
\begin{align}
    b_{t+1}^\phi\vg + c_{t+1}^\phi\phi &= \phi_{t+1} \\
                                    &= \phi_t - \eta\vv_{t+1} \\
                                    &= b_t^\phi \vg + c_t^\phi \phi - \eta\{(\mu b_t^\vv + 1 + \omega b_t^\phi)\vg + (\mu c_t^\vv + \omega c_t^\phi) \phi\} \\
                                    &= \{-\eta \mu b_t^\vv - \eta + (1-\eta\omega)b_t^\phi\}\vg + \{-\eta \mu c_t^\vv + (1-\eta\omega)c_t^\phi\}\phi.
\end{align}
\end{proof}

For the next Proposition, we record first few coefficients:
\begin{align}
    b_2^\phi &= -\eta \\
    c_2^\phi &= 1 - \eta\omega \\
    b_3^\phi &= -\eta (\mu + 2 - \eta\omega) \\
    c_3^\phi &= -\eta\mu\omega + (1-\eta\omega)^2.
\end{align}

{\bf Proposition 1, restated.} \label{prop:multistep}
{\it Multi-step estimation using SGD with momentum cannot be realized by one-step SGD with momentum consisting of different optimization hyper-parameters.}

\begin{proof}
Suppose not. We try to construct 2-step estimation with one-step by choosing different optimization hyper-parameters, and we claim that this is not possible.

Let us take 2-step estimation, \ie, we set $\vg_{2k+2}=\vg_{2k+1}=\nabla_{\phi_{2k+1}}\ls_{2k+1}(\phi_{2k+1})$ for $k=0,1,\cdots$.
By judiciously choosing optimization hyper-parameters, we try to realize this with equivalent one-step SGD with momentum.
To be specific, we attempt to find $\tilde{\mu}, \tilde{\omega}, \tilde{\eta}$ such that the sequence defined by
\begin{align}
    \tilde{\vv}_{t+1}  &\leftarrow \tilde{\mu} \tilde{\vv}_t + \tilde{\vg}_t + \tilde{\omega}\tilde{\phi}_t \\
    \tilde{\phi}_{t+1} &\leftarrow \tilde{\phi_{t}} - \tilde{\eta} \tilde{\vv}_{t+1}
\end{align}
with $\tilde{\vv_1}=0, \tilde{\phi_1}=\phi$ and $\tilde{\vg_t} = \vg_{2t-1}$ for $t=1,2,\cdots$, satisfies $\tilde{\phi_t}=\phi_{2t-1}$ for all $t\ge 1$.

For simplicity, we consider the case $\vg_t=\vg_1=\vg$ for all $t$ (for instance, we may have $\ls_t(\phi_t)=c\phi_t$ for any constant $c$). Then we can apply \lemm~\ref{lem:SGD-recur} to both $\vv_t, \phi_t$ and $\tilde{\vv}_t, \tilde{\phi}_t$, and we also borrow the notations $b_t^\vv, c_t^\vv, b_t^\phi, c_t^\phi$ and $\tilde{b}_t^\vv, \tilde{c}_t^\vv, \tilde{b}_t^\phi, \tilde{c}_t^\phi$.

Note that from $\tilde{b}_2^\phi \vg + \tilde{c}_2^\phi \phi = \tilde{\phi_2}=\phi_3=b_3^\phi \vg + c_3^\phi \phi$, it follows that $\tilde{b}_2^\phi = b_3^\phi$ and $\tilde{c}_2^\phi=c_3^\phi$.
This implies that
\begin{align}
    -\tilde{\eta} = \tilde{b}_2^\phi = b_3^\phi &= -\eta (\mu + 2 - \eta\omega) \\
    1 - \tilde{\eta}\tilde{\omega} = \tilde{c}_2^\phi = c_3^\phi &= -\eta\mu\omega + (1-\eta\omega)^2.
\end{align}
Therefore we should have
\begin{align}
    \tilde{\eta} &= \eta (\mu + 2 - \eta\omega) \\
    \tilde{\omega} &= \frac{-\eta\mu\omega + (1-\eta\omega)^2 - 1}{-\tilde{\eta}}
                   = \frac{\mu\omega + 2\omega - \eta\omega^2}{\mu + 2 - \eta\omega}.
\end{align}
Similarly, $-\tilde{\eta} (\tilde{\mu} + 2 - \tilde{\eta}\tilde{\omega}) = \tilde{b}_3^\phi = b_5^\phi$ implies that
\begin{align}
    \tilde{\mu} = -\frac{b_5^\phi}{\tilde{\eta}} + \tilde{\eta}\tilde{\omega} - 2
                = -\frac{b_5^\phi}{\eta (\mu + 2 - \eta\omega)} + \eta\mu\omega - (1-\eta\omega)^2 - 1.
\end{align}
We found all the necessary conditions for $\tilde{\mu}, \tilde{\omega}, \tilde{\eta}$, and from them we can calculate all $\tilde{b}_t^\vv, \tilde{c}_t^\vv, \tilde{b}_t^\phi, \tilde{c}_t^\phi$ given $\mu, \omega, \eta$.
Assume $\mu=0.9, \omega=0.0001, \eta=0.1$.
By the above arguments, we should have $\tilde{b}_4^\phi=b_7^\phi$ but we can check that $\tilde{b}_4^\phi \approx -1.88$ and $b_7^\phi \approx -1.78$, which is a contradiction.
\end{proof}

\section{Pre-processing and augmentation in the transfer learning}
\label{sec:preproc}
The pre-processing and augmentation are consistent with the previous works~\cite{He2015,Jang2019}.
For the small-sized inputs (32x32), the images of TinyImageNet~\cite{tinyimagenet} and STL-10~\cite{coates2011analysis} are resized to 32x32 (CIFAR-100~\cite{krizhevsky2009tiny} is already 32x32.)
The image transforms for augmentation of TinyImageNet, CIFAR-100, STL-10 are the padding of 4, randomly cropping of 32x32, and randomly horizontal flipping.
For testing, we take cropping the center of 32x32.
For the large-sized inputs (224x224), we resize the image to 256x256, randomly cropping of 224x224, randomly horizontal flipping. For testing, we take cropping the center of 224x224. We normalize the RGB channels with means and standard deviations of (0.485, 0.456, 0.406) and (0.229, 0.224, 0.225) for each channel, respectively.

\begin{table}[h!]
  \centering
  \label{tbl:transfer-datasets}
  \caption{The number of examples in each split of datasets. If there is no validation split, we randomly take a subset from training split.}
  \vspace{0.1in}
  \begin{tabular}{ccccc}
  \toprule
  Dataset & Training & Validation & Test & Total \\ 
  \midrule
  CIFAR-100~\cite{krizhevsky2009tiny}   & 45,000 & 5,000 & 10,000 & 60,000 \\
  STL-10~\cite{coates2011analysis}      & \04,500 & \0\0500 & \08,000 & 13,000 \\
  \midrule
  Bird~\cite{wah2011caltech}            & \04,994 & 1,000 &  5,794 & 11,788 \\
  Indoor~\cite{quattoni2009recognizing} & \04,360 &   1,000 &  1,340 & \06,700 \\
  Action~\cite{yao2011human}            & \03,600 & \0\0400 &  5,532 & \09,532 \\
  Dog~\cite{Khosla2011dog}              &  10,800 &   1,200 &  8,580 &  20,580 \\
  \bottomrule
  \end{tabular}
\end{table}

\section{Hyper-parameters in the transfer learning}
\label{sec:hyper-transfer}
The meta-optimizer is Adam~\cite{Kingma2014} with the fixed learning rate of 1e-3 (for 32x32 experiments) or 1e-4 (for 224x224 experiments) and the weight decay of 1e-4.
The inner-level optimizer is SGD (or Adam), with the momentum (betas) of 0.9 (0.9, 0.999) and the weight decay of 1e-4. The initial learning rate is 1e-1 (1e-3) with the cosine annealing~\cite{loshchilov2016sgdr} as follows:
\begin{align}
    \eta_t = \frac{1}{2}(1+\rm{cos}\frac{t}{T}\pi)
\end{align}
where the number of training epochs $T$ is 200. For the 32x32 experiments, the size of mini-batch is 128. For the 224x224 experiments, the size of mini-batch is 64. 
Notice that, while splitting meta-train and meta-test is a popular way in the few-shot learning, our meta-transfer learning uses the merged meta-data for the sake of efficacy as in the previous work~\cite{Jang2019}.
The other hyper-parameters are fixed between them and across multiple benchmarks.

\newpage
\section{Hyper-parameters in the few-shot learning}
\label{sec:hyper-few-shot}
We use two popular benchmark datasets, \Domni~\cite{Lake2011omniglot} and \Dmini~\cite{Ravi2017miniimagenet}.
The \Domni~dataset contains 20 images for each character where 1623 characters from 50 different alphabets.
We follow the same protocol of~\cite{Finn2017}.
The \Dmini~dataset has 100 classes with each class having 600 examples.
We follow the splits of~\cite{Ravi2017miniimagenet}, which consists of 64 training classes, 16 validation classes, and 20 test classes.

In all experiments, we use Adam~\cite{Kingma2014} with the learning rate of 0.001 for the meta-optimizer.
For the inner optimizer, we use SGD with momentum of 0.9 and weight decay of 1e-4 for MAML and Multi-step, and vanilla SGD for FO-MAML.
All models are trained and evaluated with 8 update steps, and are trained for 60,000 iterations on a single NVIDIA Tesla V100 GPU.
For \Domni, we set 0.4 for the learning rate of inner-level optimizer in all cases.
The 5-way models and 20-way models were trained with a meta batch-size of 32 tasks and 16 tasks, respectively.
Since we do not use validation split for \Domni, the results are reported by the models of final iteration.
For \Dmini, the learning rates of inner-level optimizer were chosen by grid-search using validation split among $\{0.1, 0.04, 0.01, 0.004\}$.
For MAML and Multi-step, the chosen learning rates were 0.01 for the 1-shot model, and 0.04 for the 5-shot model. For FO-MAML, they were 0.1 and 0.04 for the 1-shot and 5-shot, respectively.
Both the 1-shot and 5-shot models were trained with a meta batch-size of 4 tasks.

\section{Training time for the transfer learning}
\label{sec:supplementary_transfer_time}
\begin{figure}[h!]
    \vspace{1em}
    \begin{center}
        \includegraphics[width=.6\textwidth]{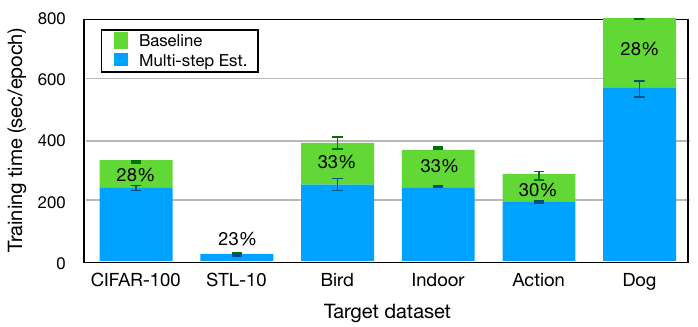}
    \end{center}
    \captionof{figure}{Training time (second per epoch) for each dataset using a Titan Xp. The labeled percentage shows the ratio of reduced time from our multi-step estimation. $\pm$ denotes the standard deviation of three randomly-initialized models.
    }
    \label{fig:transfer_time}
\end{figure}

In \fig~\ref{fig:transfer_time}, 
three-time runs show that our method significantly reduces training time across multiple datasets.  Here, meta-transfer learning boosts the performance of transfer learning~\cite{Jang2019}; however, the increased training time is one of the major drawbacks. For this reason, we argue that our method is a valuable approach to large-scale applications. In these experiments, Adam~\cite{Kingma2014} is used for inner optimization; in the paper, the training time of Bird~\cite{wah2011caltech} is slightly different since that is from a different set of experiments using the momentum SGD as an inner optimizer.

\section{Additional experiment for the transfer learning}
\label{sec:supplementary_experiment_transfer}

In this section, we observe 1) the empirical choice of $T=2$ for the number of inner steps of the transfer loss $\lstfr$ in the previous work~\cite{Jang2019}, 2) competitive performance of our multi-step estimation over the number of inner steps, and 3) the limitation of the change of a learning rate to estimate the multi-step dynamics.

The task is to transfer the knowledge of ResNet-34 pretrained on the ImageNet~\cite{russakovsky2014imagenet} to ResNet-18 for the CUB200 dataset~\cite{wah2011caltech}. We follow the experimental protocol in the paper except using Adam inner optimizer for better performance. We measure the mean accuracy with three randomly-initialized models and the training time (second) for an epoch with their standard deviations.

\begin{figure}[h!]
    \vspace{1em}
    \begin{center}
        \includegraphics[width=.6\textwidth]{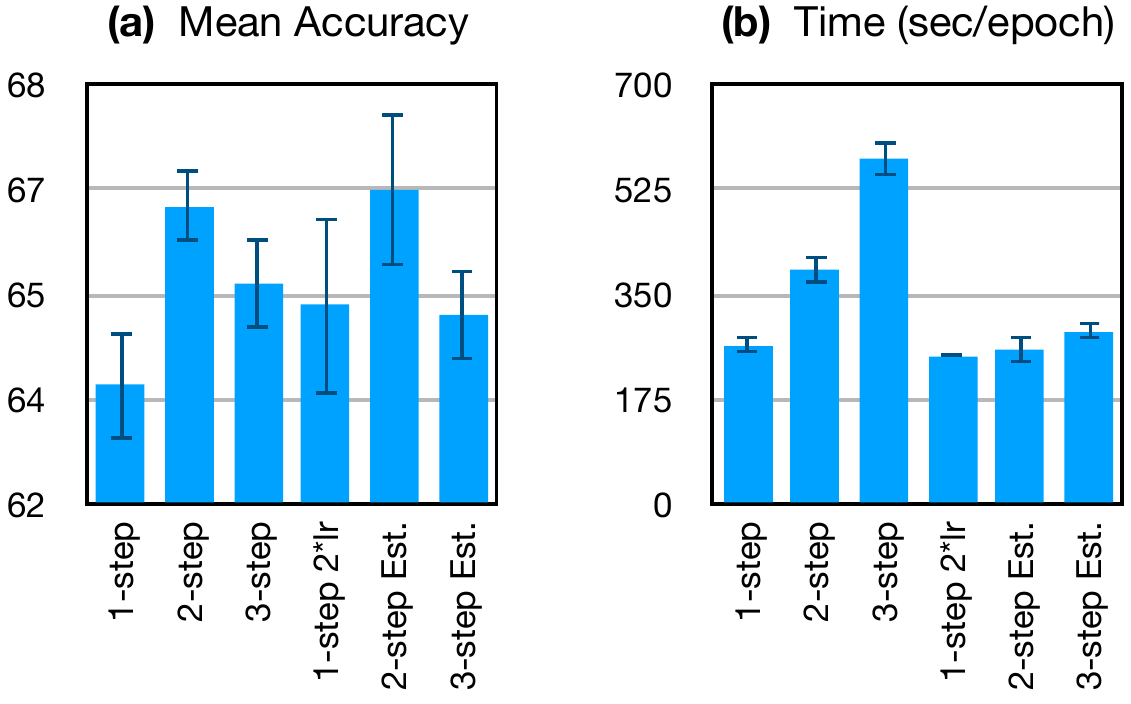}
    \end{center}
    \captionof{figure}{Mean accuracy and training time of the transfer learning for the \Dcub~dataset~\cite{wah2011caltech}. 
    }
    \label{fig:cub200_comp}
\end{figure}

In \fig~\ref{fig:cub200_comp}, we empirically confirm the choice of $T=2$ for its performance (\fig~\ref{fig:cub200_comp}a) considering the trade-off with the training time depending on the number of inner steps (\fig~\ref{fig:cub200_comp}b). The training time linearly increases as the number of inner steps increases and it is critical for high-dimensional tasks.
Our multi-step estimation methods, \textit{\{2,3\}-step Est.}, competitively perform or even outperform their counterparts, \textit{\{2,3\}-step}.
When we use a doubled learning rate, \textit{1-step 2*lr}, to estimate \textit{2-step} with a single step, it deteriorates the performance compared with our \textit{2-step Est.}, since it simply ignores the dynamics of inner-level.
With $T=0$, the performance plummets to $48.91\pm0.83$, and the training time per epoch is $121.3\pm4.97$ seconds.

\end{document}